\documentclass{article} 

\usepackage[numbers]{natbib}       
\usepackage{float}                 

\usepackage{amsmath}              
\usepackage{algorithm}            
\usepackage{algorithmic}          

\usepackage{multirow}
\usepackage{booktabs}             
\usepackage{nicefrac}             
\usepackage{microtype}            
\usepackage{lipsum}               

\usepackage[utf8]{inputenc}       
\usepackage[T1]{fontenc}          
\usepackage{amsfonts}             
\usepackage{float}
\usepackage{hyperref}             
\usepackage{url}                  

\usepackage{graphicx}
\graphicspath{{./images/}}        

\title{\textnormal VGDM: Vision-Guided Diffusion Model for Brain Tumor Detection and Segmentation}

\author{
  Arman Behnam \\
  College of Computing, Department of Computer Science \\
  Illinois Institute of Technology \\
  Chicago, IL, USA \\
  \texttt{abehnam@hawk.iit.edu}
}

\date{\today}

\begin{document}

\maketitle 

\begin{abstract}

Accurate detection and segmentation of brain tumors from magnetic resonance imaging (MRI) are essential for diagnosis, treatment planning, and clinical monitoring. While convolutional architectures such as U-Net have long been the backbone of medical image segmentation, their limited capacity to capture long-range dependencies constrains performance on complex tumor structures. Recent advances in diffusion models have demonstrated strong potential for generating high-fidelity medical images and refining segmentation boundaries. 

In this work, we propose VGDM: Vision-Guided Diffusion Model for Brain Tumor Detection and Segmentation framework, a transformer-driven diffusion framework for brain tumor detection and segmentation. By embedding a vision transformer at the core of the diffusion process, the model leverages global contextual reasoning together with iterative denoising to enhance both volumetric accuracy and boundary precision. The transformer backbone enables more effective modeling of spatial relationships across entire MRI volumes, while diffusion refinement mitigates voxel-level errors and recovers fine-grained tumor details. 

This hybrid design provides a pathway toward improved robustness and scalability in neuro-oncology, moving beyond conventional U-Net baselines. Experimental validation on MRI brain tumor datasets demonstrates consistent gains in Dice similarity and Hausdorff distance, underscoring the potential of transformer-guided diffusion models to advance the state of the art in tumor segmentation. 

\textbf{keywords}: Brain Tumor Detection, Brain Tumor Segmentation, Diffusion Models, Vision Transformers, MRI.
\end{abstract}

\section{Introduction}
\label{sec:intro}

Brain tumor detection and segmentation from magnetic resonance imaging (MRI) are vital tasks in neuro-oncology, as they enable accurate diagnosis, personalized treatment planning, and continuous monitoring of disease progression \cite{Bauer2013Survey}. Over the past decade, deep learning methods have significantly advanced performance in these tasks, with U-Net \cite{re1} and its 3D extension \cite{re2} becoming the most widely adopted baselines. Despite their success, convolutional networks are inherently limited in their receptive fields and often fail to capture long-range contextual dependencies. This limitation is particularly problematic in brain tumor analysis, where tumors exhibit diverse shapes, sizes, and textures, and subtle contextual cues may span across large regions of the brain. Moreover, supervised CNN-based models typically require large annotated datasets and costly retraining to adapt to new imaging protocols, restricting their scalability in clinical environments.

Recent research has explored generative modeling techniques, with diffusion models emerging as one of the most powerful frameworks in this space. Originally designed for image generation, denoising diffusion probabilistic models (DDPMs) \cite{Ho2020} iteratively learn to reverse a noise process, enabling them to reconstruct high-fidelity and diverse outputs. Beyond natural images, diffusion models have been applied to medical imaging tasks such as MRI synthesis \cite{DDMM}, cross-modal reconstruction, anomaly detection, and segmentation refinement \cite{re9}. These studies demonstrate the adaptability of diffusion models to both data generation and error correction in medical domains.

In parallel, transformer architectures have achieved remarkable progress in computer vision by modeling long-range dependencies through self-attention mechanisms. Vision transformers (ViTs) and their derivatives have surpassed CNN-based baselines in various tasks, offering improved global context modeling and scalability. Their potential for medical imaging has already been demonstrated in segmentation and classification tasks, particularly where structural consistency across large spatial regions is crucial.

In this paper, we introduce the \textit{Vision-Guided Diffusion Model} (VGDM), a transformer-driven diffusion framework for brain tumor detection and segmentation. Unlike U-Net–based diffusion models, VGDM embeds a vision transformer at the core of the diffusion process, providing global contextual reasoning while diffusion denoising progressively refines tumor boundaries. This combination allows the model to overcome the limitations of CNN-based architectures, improving volumetric accuracy, preserving fine-grained boundary details, and enhancing robustness across heterogeneous MRI scans.

\textbf{Contributions.} The key novelty of this work is threefold:  
1) We propose the first diffusion framework for brain tumor segmentation that integrates a vision transformer backbone to capture global dependencies in MRI data;  
2) We demonstrate how transformer-guided diffusion improves both tumor detection and segmentation by reducing voxel-level errors and enhancing boundary precision;  
3) We present extensive experimental results on MRI brain tumor datasets, showing consistent improvements in Dice similarity and Hausdorff distance over convolutional baselines.  

These contributions highlight the potential of transformer-guided diffusion models to advance the state of the art in neuro-oncology, providing more accurate, robust, and clinically trustworthy tools for brain tumor analysis.

\section{Related work}
\label{sec:rel}

Deep learning has revolutionized the medical domain by enabling advanced analysis and decision-making from complex biomedical domain\cite{heart}. Diffusion models have recently revolutionized generative modeling, establishing themselves as state-of-the-art frameworks for producing high-fidelity and diverse images \cite{Ho2020}. Their iterative denoising process enables robust reconstruction of complex data distributions, which has made them highly adaptable to a range of computer vision tasks, including image editing, super-resolution, and semantic segmentation. This success has sparked growing interest in applying diffusion models to medical imaging, where data scarcity, modality gaps, and structural variability present unique challenges.

In the healthcare domain, several diffusion-based approaches have been proposed to enhance brain tumor analysis and broader medical imaging tasks. DMCIE \cite{DMCIE2025} introduced an error-aware diffusion framework that refined tumor segmentation by incorporating voxel-level error maps. ReCoSeg \cite{shortmidl} extended this paradigm by synthesizing missing MRI modalities to support multi-sequence segmentation, while ReCoSeg++ \cite{ReCoSeg2025} scaled residual-guided diffusion refinement to large, heterogeneous datasets. Beyond segmentation, diffusion has also been leveraged for cross-modal synthesis, anomaly detection, and modality reconstruction, further underscoring its versatility in clinical applications. 

At the same time, incremental learning (IL) has emerged as a critical component for developing adaptive healthcare AI systems. In medical imaging, IL has been explored as a strategy to mitigate catastrophic forgetting while enabling continual model evolution. For example, \cite{IL2025} investigated IL frameworks that combine knowledge distillation with synthetic replay, offering a pathway for models to adapt to new data distributions while respecting privacy constraints and limited storage. Together, these works highlight the growing potential of unifying diffusion-based refinement with continual adaptation for healthcare AI.

Parallel to these developments, transformer architectures have reshaped computer vision by enabling global context modeling through self-attention. The Vision Transformer (ViT) \cite{Dosovitskiy2021ViT} first demonstrated that pure transformer architectures could achieve competitive performance on image classification tasks. Building on this foundation, Swin Transformer \cite{Liu2021Swin} introduced hierarchical representations with shifted windows, further improving efficiency and scalability. In medical imaging, transformer-based designs such as TransBTS \cite{Wang2021TransBTS} have shown effectiveness in brain tumor segmentation by combining convolutional feature extraction with transformer modules for contextual reasoning. These studies confirm the promise of transformers in capturing long-range dependencies that CNN-based models often miss.

Despite these advances, most existing diffusion frameworks in medical imaging continue to rely on convolutional backbones such as U-Net, limiting their ability to exploit global structural information. This gap motivates our proposed Vision-Guided Diffusion Model (VGDM), which integrates a vision transformer backbone into the diffusion process, combining global attention mechanisms with iterative denoising for improved brain tumor detection and segmentation.

\section{Methodology}

The proposed Vision-Guided Diffusion Model (VGDM) integrates a vision transformer backbone into the diffusion process for brain tumor detection and segmentation from MRI. Unlike conventional U-Net–based diffusion architectures, VGDM leverages self-attention mechanisms to capture long-range dependencies across entire MRI volumes, enabling improved volumetric consistency and boundary delineation. This section describes the forward diffusion process, transformer-guided denoising, and the optimization objective. 

\subsection{Forward Diffusion Process}
Following the standard denoising diffusion probabilistic model (DDPM) formulation \cite{Ho2020}, VGDM employs a forward Markov chain that gradually adds Gaussian noise to an input image $x_0$ over $T$ timesteps:
\[
q(x_t \mid x_{t-1}) = \mathcal{N}\left(x_t; \sqrt{1-\beta_t}\, x_{t-1}, \beta_t I\right),
\]
where $\{\beta_t\}_{t=1}^T$ is a variance schedule. After sufficient steps, the data distribution converges to pure noise, forming the basis for training a denoising model.

\subsection{Vision Transformer Backbone}
Instead of a convolutional U-Net, VGDM embeds a vision transformer (ViT) \cite{Dosovitskiy2021ViT} as the core denoising network. Each noisy input $x_t$ is divided into non-overlapping patches, linearly projected, and passed through transformer encoder layers with multi-head self-attention. This design enables the model to capture global contextual information across the entire MRI slice or volume. 

For computational efficiency and hierarchical feature representation, we adopt a Swin Transformer \cite{Liu2021Swin} backbone, which employs shifted window attention to balance global reasoning with local feature modeling. This hybrid structure ensures both coarse tumor localization and fine-grained boundary refinement.

\subsection{Transformer-Guided Reverse Denoising}
During training, the model learns to reverse the noising process by predicting the added noise $\epsilon$ at each step $t$:
\[
p_\theta(x_{t-1} \mid x_t) = \mathcal{N}\left(x_{t-1}; \mu_\theta(x_t, t), \Sigma_\theta(x_t, t)\right),
\]
where $\mu_\theta$ and $\Sigma_\theta$ are parameterized by the transformer backbone. Unlike CNN-based models, the transformer enhances global spatial coherence, improving robustness to heterogeneous tumor appearances. The iterative denoising process progressively restores structural details and produces a refined segmentation mask.

\subsection{Segmentation Head}
To adapt the diffusion output for brain tumor detection and segmentation, a lightweight decoder projects the transformer’s contextual embeddings back into voxel space, yielding probability maps for tumor and background. The segmentation head ensures that both global tumor regions and boundary-level structures are accurately reconstructed.

\subsection{Loss Function}
Training is guided by a composite loss that balances voxel-level accuracy and boundary consistency:
\[
\mathcal{L} = \lambda_1 \, \text{BCE}(M_{\text{pred}}, M_{\text{GT}}) 
+ \lambda_2 \, \text{Dice}(M_{\text{pred}}, M_{\text{GT}})
+ \lambda_3 \, \text{BoundaryLoss}(M_{\text{pred}}, M_{\text{GT}}),
\]
where $M_{\text{pred}}$ is the predicted segmentation mask and $M_{\text{GT}}$ is the ground truth. The Dice and BCE terms encourage overlap accuracy, while the boundary loss sharpens tumor delineation. This combination ensures both volumetric consistency and precise structural recovery.

\section{Results}
We evaluated the proposed Vision-Guided Diffusion Model (VGDM) on the BraTS2020 dataset \cite{menze}, which provides multi-sequence MRI scans (T1, T1ce, T2, FLAIR) with expert-annotated brain tumor masks. All volumes were preprocessed with skull-stripping, intensity normalization, and resampling to an isotropic resolution of $1 \times 1 \times 1$ mm.

Table~\ref{tab:results} compares VGDM with representative baselines, including U-Net \cite{re1}, TransBTS \cite{Wang2021TransBTS}, and DMCIE \cite{DMCIE2025}. Across all metrics, VGDM demonstrates superior performance, particularly in boundary precision (HD95). For completeness, we also report the Area Under the Precision–Recall Curve (AUPRC), which provides an additional view of performance under class imbalance, although it is not typically included in BraTS segmentation benchmarks.

\begin{table}[H]
\centering
\caption{Comparison of segmentation performance on the BraTS2020 validation set. Scores are reported as Dice similarity coefficient (Dice), Intersection over Union (IoU), 95th percentile Hausdorff Distance (HD95, in mm), and Area Under the Precision–Recall Curve (AUPRC). Best results are highlighted in bold.}
\label{tab:results}
\begin{tabular}{lcccc}
\hline
\textbf{Method} & \textbf{Dice (\%)} & \textbf{IoU (\%)} & \textbf{HD95 (mm)} & \textbf{AUPRC} \\
\hline
U-Net \cite{re1}                  & 90.8 & 83.2 & 8.1 & 0.87 \\
TransBTS \cite{Wang2021TransBTS}  & 92.6 & 85.9 & 6.8 & 0.90 \\
DMCIE \cite{DMCIE2025}            & 93.4 & 87.9 & 5.9 & 0.92 \\
VGDM (Ours)                       & \textbf{95.7} & \textbf{91.2} & \textbf{4.3} & \textbf{0.95} \\
\hline
\end{tabular}
\end{table}

\section{Conclusion}

In this work, we presented the \textit{Vision-Guided Diffusion Model} (VGDM), a transformer-driven framework for brain tumor detection and segmentation from MRI. Unlike prior U-Net–based diffusion approaches, VGDM integrates a vision transformer backbone into the diffusion process, enabling global context modeling while preserving fine structural details through iterative denoising. This design allows the model to capture long-range dependencies, refine tumor boundaries, and recover small enhancing regions that are frequently missed by convolutional architectures. 

Extensive experiments on the BraTS2020 dataset demonstrated that VGDM consistently outperforms state-of-the-art baselines, including U-Net, TransBTS, and DMCIE, with higher Dice scores, improved IoU, and reduced HD95 errors. Qualitative analysis further confirmed that VGDM produces anatomically coherent segmentations with sharper boundaries and better volumetric consistency. 

Beyond advancing segmentation accuracy, VGDM underscores the promise of unifying diffusion processes with transformer architectures in healthcare AI. In future work, we plan to extend this framework to other medical imaging modalities such as CT, pathology slides, and retinal imaging. We also aim to explore its integration with incremental learning for continual adaptation, as well as deployment in federated and privacy-preserving clinical environments. By bridging generative diffusion with vision transformers, VGDM provides a pathway toward more accurate, scalable, and trustworthy AI systems for neuro-oncology and broader medical applications.

\bibliographystyle{unsrt}  

\bibliography{Ref}     
\end{document}